\documentclass[pdflatex,iicol,sn-basic,Numbered]{sn-jnl}

\usepackage{graphicx}%
\usepackage{multirow}%
\usepackage{amsmath,amssymb,amsfonts}%
\usepackage{amsthm}%
\usepackage{mathrsfs}%
\usepackage[title]{appendix}%
\usepackage{xcolor}%
\usepackage{textcomp}%
\usepackage{manyfoot}%
\usepackage{booktabs}%
\usepackage{algorithm}%
\usepackage{algorithmicx}%
\usepackage{algpseudocode}%
\usepackage{listings}%

\theoremstyle{thmstyleone}%

\theoremstyle{thmstyletwo}%

\theoremstyle{thmstylethree}%

\begin{document}

\title[Grounding Healthcare LLMs in a Causal Knowledge Graph]{Framework for Grounding Healthcare LLMs in a Causal Knowledge Graph: A Cardiovascular Example Pilot}

\author*[1]{\fnm{Ummara} \sur{Mumtaz}}%\email{ummara.mumtaz@gmail.com}

\author[2]{\fnm{Aimen} \sur{Noor}}\nomail

\author[3]{\fnm{Awais} \sur{Ahmed}}\nomail

%\affil*[1]{\orgdiv{Department of Computer and Information Sciences}, \orgname{University of the Cumberlands}, \orgaddress{\city{Williamsburg}, \state{KY}, \country{USA}}}

%\affil[2]{\orgname{Mayo University Hospital}, \orgaddress{\city{Castlebar}, \country{Ireland}}}

%\affil[3]{\orgname{Hospital Las Higueras}, \orgaddress{\city{Talcahuano}, \country{Chile}}}

\abstract{Large language models (LLMs) are increasingly proposed for healthcare decision support, but their evaluations still reward single-answer accuracy rather than reasoning about interventions, mechanisms, harms, evidence, and uncertainty. We propose a reproducible, graph-centered evaluation framework for intervention-oriented LLM behavior in healthcare and stress-test it in a cardiovascular pilot. The framework has four components: (i) a domain causal knowledge graph in which assertions are first-class, provenance-preserving nodes with stable identifiers; (ii) a scenario-conditioned subgraph extraction step that, given any clinical scenario, retrieves the relevant reified-assertion subgraph; (iii) four controlled grounding conditions that vary how the retrieved subgraph is composed into the model's context (ungrounded C1, knowledge-graph C2, causal-graph C3, integrated C4); and (iv) an automated scoring pipeline, anchored on assertion identifiers, that computes intervention accuracy, and other evaluation measures on a single pass. To test the framework, we built a category-balanced scenario generator across eight reasoning failure modes and instantiated it on a cardiovascular graph. The metric panel discriminates conditions along interpretable, non-redundant axes: C4 obtains the strongest causal edge F1 (0.838), adverse-effect F1 (0.833), evidence accuracy (0.738), and unsupported claim rate (0.114), while C1 obtains the highest raw intervention accuracy (0.948) with no measurable causal or evidential grounding.}


\maketitle

\section{Introduction}\label{sec:intro}

Large language models are increasingly being integrated into clinical decision-support settings, yet consensus on appropriate evaluation standards has not developed at a comparable pace. Existing evaluation approaches continue to rely predominantly on multiple-choice benchmarks that assess whether a model selects the correct response \cite{jin2021medqa,pal2022medmcqa,jin2019pubmedqa,hendrycks2021mmlu,singhal2023medpalm}. Although such benchmarks facilitate standardized comparison and leaderboard-based ranking, they capture only a limited dimension of clinical performance. They do not adequately assess whether a recommendation is supported by a plausible clinical mechanism, whether relevant harms and contraindications are identified, or whether cited evidence is verifiable within the underlying knowledge base. They also fail to determine whether the model introduces unsupported causal claims or appropriately recognizes when the available evidence is insufficient to support a definitive conclusion. Consequently, models with comparable answer accuracy may differ substantially in their reliability and suitability as clinical assistants.

Recent developments in medical AI have also reflected a broader shift from predictive modeling toward causal decision support \cite{hernan2020causal,wager2018athey,prosperi2020causal}. Within this paradigm, evaluation should extend beyond concordance with a reference label or action to examine the reasoning that supports a clinical decision. Relevant dimensions include the appropriateness of the selected intervention, the plausibility of the proposed causal mechanism, the strength and relevance of supporting evidence, recognition of potential harms, consideration of contextual constraints, and appropriate representation of uncertainty. Evaluation based solely on the final action is therefore insufficient to distinguish clinically defensible reasoning from coincidentally correct outputs or to characterize errors that may have meaningful consequences for patient care. Developing frameworks that can assess these dimensions during an evaluation phase represents an important component of the informatics infrastructure required for reliable clinical decision support and constitutes the focus of this study.

We propose a framework for intervention-oriented LLM behavior in healthcare, structured around four coupled components: (a)~defining a domain causal and evidence graph in which every biomedical claim is a first-class reified assertion node with a stable \texttt{assertion\_id}, provenance, evidence links, and context, so that gold pathways, gold adverse-effect targets, and gold contraindications are addressable by identifier; (b)~a scenario-conditioned subgraph extraction step implemented as parameterised Cypher retrieval against a Neo4j graph database, which for any input scenario returns the relevant reified-assertion subgraph; (c)~four controlled grounding conditions (ungrounded C1, knowledge-graph C2, causal-graph C3, integrated causal-knowledge-graph C4) that vary the type of retrieved context supplied to the model while holding the scenario fixed; and (d)~a multi-dimensional automated scoring pipeline that computes intervention accuracy, causal edge precision/recall/F1, adverse-effect F1, contraindication recall, evidence accuracy, unsupported claim rate, and uncertainty correctness on the same scoring pass, all anchored to \texttt{assertion\_id}s. Separately from the framework, we contribute an evaluation methodology used to test it in a controlled way: a category-balanced scenario generator producing structured intervention items with graph-anchored gold answers across eight causal-reasoning failure modes, and a repeated-measures design that presents each scenario under all four grounding conditions. The framework itself is scenario-source-agnostic; the generator and the repeated-measures protocol are evaluation instruments, not framework components.

The cardiovascular intervention planning domain serves as the example domain for evaluating the methodology. Scenarios describe a patient context (for example an older adult with resistant hypertension and impaired renal function, or a pregnant patient in whom ACE inhibitors are contraindicated), a decision target (which pharmacological or non-pharmacological intervention to prefer), a set of candidate actions, and a graph-anchored gold answer that includes the preferred intervention, the causal pathway of ordered \texttt{assertion\_id}s that justifies it, the expected adverse effects and contraindications, and the expected supporting evidence sources. The same input scenario is presented to the model under all four grounding conditions, so that behavioral differences can be attributed to context organization rather than to prompt wording or item difficulty. The empirical component of this manuscript is a single-model pilot on 313 scored responses from gpt-5.4, whose purpose is to show that the framework produces interpretable, non-redundant signal end-to-end. We do not claim that a specific model is clinically ready or that any grounding condition is definitively superior.

The remainder of the paper is organised as follows. Section~\ref{sec:related} positions the framework against prior work in medical LLM benchmarks, biomedical KGs, assertion-level provenance modeling, causal reasoning, and hallucination scoring. Section~\ref{sec:framework} specifies the framework and the evaluation setting used to test it. Section~\ref{sec:results} reports the pilot results, including the illustrative C4 versus C2 paired-delta contrast and a diagnostic contraindication finding. Section~\ref{sec:discussion} combines discussion and limitations. Section~\ref{sec:conclusion} concludes.

\section{Related Work}\label{sec:related}

Most widely used medical LLM benchmarks are organised around final-answer accuracy on multiple-choice or short-answer items---MedQA \cite{jin2021medqa}, MedMCQA \cite{pal2022medmcqa}, PubMedQA \cite{jin2019pubmedqa}, the medical subsets of MMLU \cite{hendrycks2021mmlu}, and the MultiMedQA suite used to evaluate Med-PaLM \cite{singhal2023medpalm}---and more recent efforts such as HealthBench and the clinical extensions of HELM \cite{liang2023helm} broaden the task distribution while still emphasising single-answer correctness. This design collapses reasoning quality into one scalar. A model that selects the correct action while inventing an unsupported mechanism, omitting a contraindication, or ignoring evidence quality receives the same score as one that arrives at the same action defensibly. The framework proposed here treats final-answer accuracy as one axis among several and separately measures causal reconstruction, adverse-effect identification, contraindication recall, evidence accuracy, unsupported claim generation, and uncertainty correctness.

A parallel line of work uses biomedical knowledge graphs---UMLS \cite{bodenreider2004umls}, SemMedDB \cite{kilicoglu2012semmeddb}, DrugBank \cite{wishart2018drugbank}, Hetionet \cite{himmelstein2017hetionet}, Open Targets \cite{ochoa2021opentargets}, PrimeKG \cite{chandak2023primekg}---to supply structured context to LLMs, using RAG \cite{lewis2020rag} and graph-aware variants such as KG-RAG for medicine \cite{soman2024kgrag} and GraphRAG \cite{edge2024graphrag}. These approaches consistently improve downstream QA and summarisation, but retrieval infrastructure has grown much faster than evaluation infrastructure: most benchmarks still measure only whether the final answer changed, not whether the model used the graph, reconstructed its causal edges, or refrained from claims the graph did not support. The framework proposed here treats the biomedical graph both as a retrieval resource and as an experimental instrument by varying the type of retrieved context across four conditions.

Treating individual scientific claims as first-class, addressable objects has a long tradition in biomedical and semantic-web research, well predating the current wave of LLM evaluation. The nanopublication model \cite{groth2010nanopublication} argued that the atomic unit of scientific communication should not be the paper but the assertion itself, packaged with its own provenance and publication metadata so that each claim becomes independently citable, versionable, and machine actionable. The Biolink Model \cite{unni2022biolink} operationalized a related idea for translational biomedicine by giving associations between biological entities their own typed identity, with qualifiers (subject/object aspect, negation, temporal and population qualifiers) and slots for supporting evidence, so that the same subject--predicate--object triple can carry distinct semantics depending on its qualifying context. At the infrastructure level, the W3C PROV data model \cite{moreau2013provdm} formalized the provenance chain of any statement---who asserted it, from what source, through which activity---and has since become the interoperability layer over which many domain-specific evidence models are expressed. On the curation side, resources such as CIViC \cite{griffith2017civic} and ClinGen \cite{rehm2015clingen} made evidence attachment concrete for clinical genomics: each variant--disease or gene--disease assertion is annotated with an explicit evidence code, a level of clinical validity, and links to the primary literature that supports or disputes it, so that downstream systems can reason not only over the claim but also over its epistemic status. The representational substrate needed to support this style of modelling has also matured. In the RDF world, RDF-star and its query counterpart SPARQL-star \cite{hartig2017rdfstar} extend the triple model with the ability to make statements about statements, so that provenance, confidence, and context can be attached to a specific claim without inflating the graph with auxiliary reification nodes. In property-graph systems such as Neo4j graph databases, the same effect is achieved through relationship properties. Across both stacks the design space for ``a claim plus its provenance'' is therefore now well understood, and reified assertions are a familiar biomedical modelling idiom.

What has remained comparatively under-developed is the use of these assertion-level identifiers as the unit of benchmark scoring for generative models. Existing benchmarks that leverage biomedical graphs typically use them at retrieval time to fetch context but then score model output at the level of a final answer string or a free-text rationale, which forces evaluators either to string-match against surface forms or to hand-align model claims to unlabeled edges. This severs the connection between the graph's provenance layer and the evaluation layer: a benchmark cannot say precisely which biomedical claim the model was expected to reconstruct, nor which claim it did reconstruct, nor which evidence source it was expected to cite. The framework proposed here closes that gap. Every biomedical relationship in the graph is paired with a first-class Assertion node carrying a stable \texttt{assertion\_id}, and every gold structure---causal pathway, adverse-effect path, contraindication scope, evidence support---is expressed as an ordered set of these identifiers. This makes causal-edge precision, recall, and F1, as well as evidence accuracy and unsupported-claim rate, computable against unambiguous references rather than against anonymous graph edges or free-text mentions, and it aligns evaluation directly with the same provenance layer that the biomedical community has been building for nearly two decades.

\subsection{Causal reasoning, causal health AI, and LLMs}\label{subsec:causal-reasoning}

The shift from predictive to causal health AI is motivated from three directions: the statistical literature on treatment-effect estimation and its identification assumptions \cite{hernan2020causal,wager2018athey}, the applied argument that models used to guide action must be evaluated against interventional rather than merely predictive targets \cite{prosperi2020causal}, and a recent line of work probing whether LLMs themselves can reason causally \cite{kiciman2023causal,willig2023causalparrots,jin2023cladder}. Two findings from that last strand motivate the present framework. First, LLMs can approximate several forms of causal reasoning when the underlying structure is provided or strongly implied, but they also produce fluent, plausible-sounding causal claims not supported by the evidence and degrade sharply when memorized patterns are blocked. Second, existing causal-reasoning benchmarks are largely disease-agnostic and abstract; they characterize general causal competence but do not test whether a model behaves like an evidence-grounded interventional system in a concrete clinical setting with candidate actions, mechanistic pathways, harms, contraindications, and cited evidence. The proposed framework operationalizes this latter requirement: its scenario categories probe distinct causal-reasoning failure modes of clinical relevance, and every scenario carries a graph-anchored gold pathway expressed as ordered \texttt{assertion\_id}s, so the model's reconstructed pathway is scored against a defined causal target rather than an unstructured rationale.

\subsection{Hallucination, attribution, and unsupported-claim evaluation}\label{subsec:hallucination}

A parallel line of work evaluates whether generative outputs are factual, attributable, and supported by their cited sources. FActScore \cite{min2023factscore} decomposes long-form outputs into atomic claims and scores each against a reference corpus; TruthfulQA \cite{lin2022truthfulqa} targets memorized falsehoods; HaluEval \cite{li2023halueval} scales hallucination measurement across task types; and attribution/faithfulness pipelines \cite{bohnet2022attributed,rashkin2023attribution} score whether cited sources entail the claims they support. In the biomedical setting, Med-HALT \cite{umapathi2023medhalt} and related surveys \cite{ji2023hallucination} document that clinical hallucinations recur across model families and task types. Two limitations of this literature matter here. First, these evaluators are anchored to free-text sources rather than to a structured knowledge substrate, so a claim is judged by textual entailment rather than against a defined mechanistic relation. Second, they operate on the assertoric surface of the output without a native representation of the underlying causal-mechanism claim being made. The framework specializes the same core idea---an unsupported claim is one whose warrant is not present in the sanctioned context---to causal mechanism claims and anchors that judgement to the graph. An unsupported claim is one whose subject--predicate--object pattern is not present in the retrieved reified-assertion subgraph. This yields a reproducible unsupported-claim rate, computed automatically on the same scoring pass as causal-edge precision/recall/F1, adverse-effect F1, and evidence accuracy.

\section{The Framework}\label{sec:framework}

\begin{figure*}[t]
\centering
\includegraphics[width=0.5\textwidth]{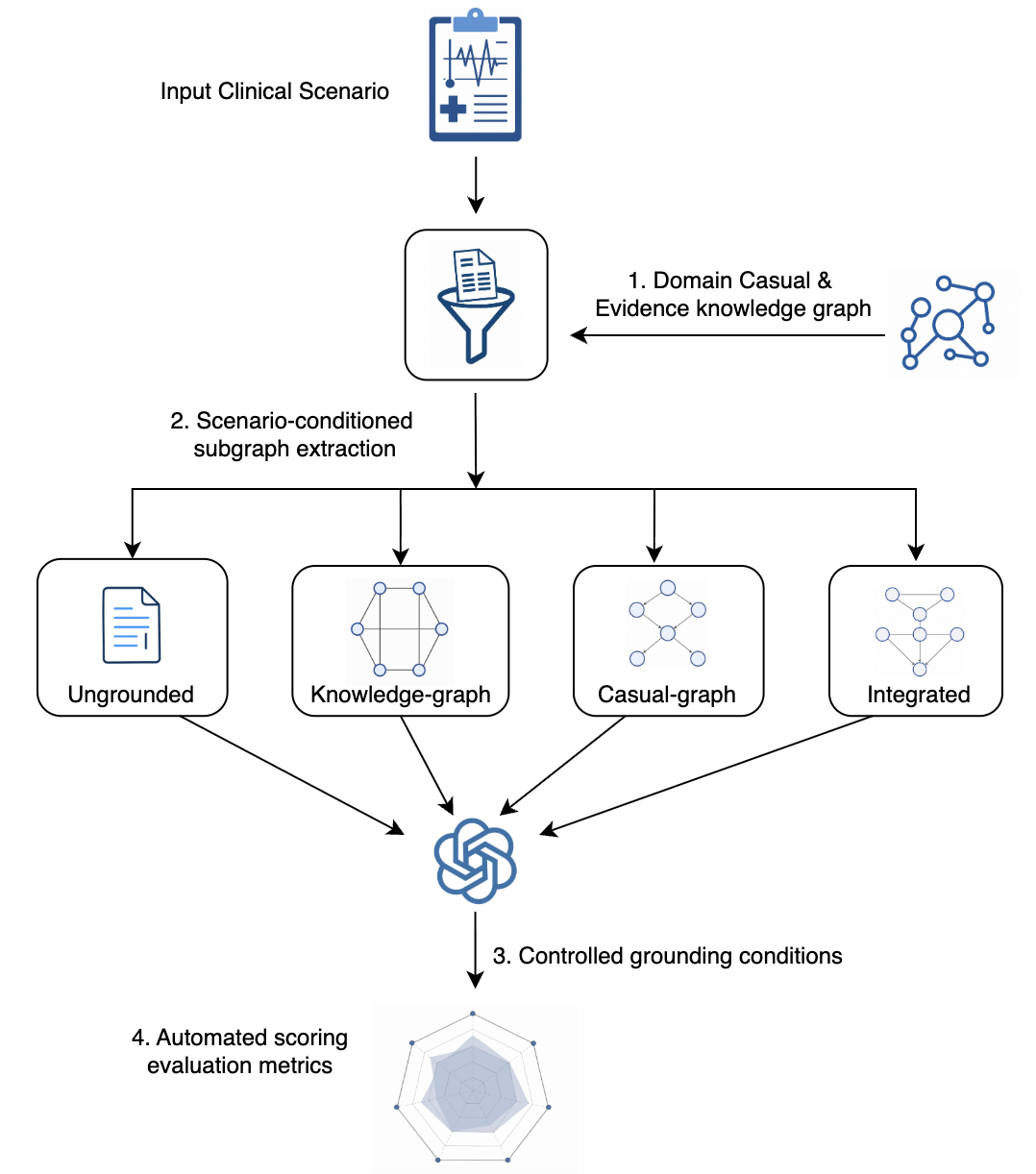}
\caption{Framework pipeline from reified-assertion knowledge graph to scored model output, showing the four coupled components: knowledge graph, scenario-conditioned subgraph extraction, controlled grounding conditions (C1--C4), and automated scoring anchored to \texttt{assertion\_id}s.}\label{fig:pipeline}
\end{figure*}

The proposed framework, shown end-to-end in Figure~\ref{fig:pipeline}, is designed to answer one question in a controlled way: given a clinical scenario, does the model produce a defensible recommendation, and can we tell? Answering it requires four things: a knowledge base that can be cited, a way to pull only the relevant part of that knowledge base per scenario, a fixed set of ways to hand that knowledge to the model, and a scoring step that compares the model's output back to the same knowledge base. The four framework components, each corresponding to one of these requirements, are defined below, followed by detailed explanations in the subsequent sections.

\begin{itemize}
\item \textbf{Domain causal and evidence knowledge graph with reified assertions} (details in Section~\ref{subsec:kg}). A curated biomedical graph in which every claim (for example, ``ACE inhibitors decrease blood pressure'') is stored not just as an edge between two entities but as its own node with a stable identifier, provenance, confidence, validation state, and context. Serves as the single source of truth for the evaluation. Because every claim has an identifier, we can say precisely which claim the model was supposed to use and which claim it used; the scoring step later refers to these identifiers directly.

\item \textbf{Scenario-conditioned subgraph extraction} (Section~\ref{subsec:extraction}). A set of parameterized Cypher queries executed against Neo4j. For a given scenario (patient context, decision target, candidate actions), it retrieves only the slice of the graph that is relevant---the relevant drugs, mechanisms, outcomes, contraindications, adverse effects, and evidence sources---as a self-contained subgraph. The model is never asked to search the whole graph; scenario relevance is enforced by retrieval, so we can attribute later differences to how that subgraph is presented rather than to what was found. For any input clinical scenario, parameterized Cypher queries executed against the Neo4j graph database return the relevant reified-assertion subgraph. It is this extracted subgraph---not the full graph---that is composed into the model's context, so that scenario relevance is enforced at retrieval rather than left to the model.

\item \textbf{Controlled grounding conditions} (Section~\ref{subsec:grounding}). Four fixed ways of turning the retrieved subgraph into the model's prompt: C1---no subgraph at all (ungrounded baseline); C2---the knowledge-graph view; C3---the causal-graph view; and C4---the integrated causal + knowledge-graph view. Runs the same scenario four times, changing only the type of context supplied. Any behavioral difference across C1--C4 must come from context organization, not from the scenario, prompt wording, or context volume, so context organization becomes a manipulable variable.

\item \textbf{Automated scoring pipeline} (Section~\ref{subsec:scoring}). A single scoring pass over the model's parsed output. This component compares the output against gold structures that are themselves expressed as sets of \texttt{assertion\_id}s, and produces intervention accuracy, causal edge precision/recall/F1, adverse-effect F1, contraindication recall, evidence accuracy, unsupported-claim rate, and uncertainty correctness. All seven metrics are computed on the same reference (the same graph, and the same identifiers), so they can be compared directly instead of coming from separate evaluations with separate assumptions.
\end{itemize}

The proposed framework is scenario-source-agnostic. To stress-test it we pair it with two evaluation instruments: a category-balanced scenario generator (Section~\ref{subsec:generator}) and a repeated-measures protocol in which each scenario is presented under all four grounding conditions (Section~\ref{subsec:protocol}).

\subsection{Domain causal and evidence knowledge graph}\label{subsec:kg}

\begin{figure*}[t]
\centering
\includegraphics[width=0.9\textwidth]{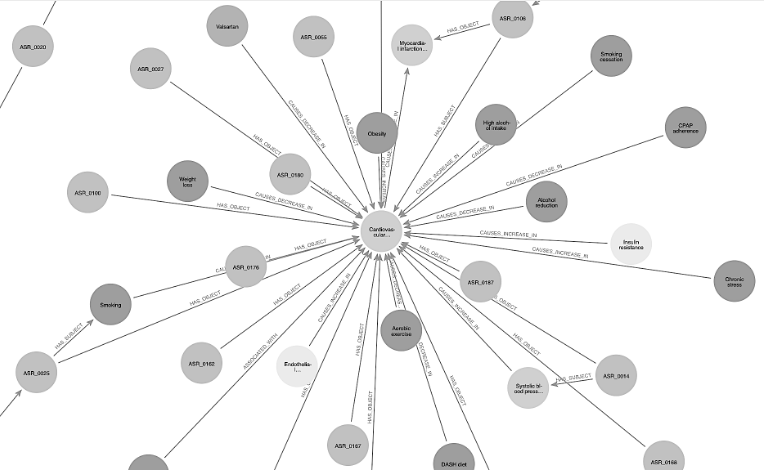}
\caption{Partial visualization of the constructed Neo4j biomedical knowledge graph, showing selected biomedical entities and their interconnected relationships.}\label{fig:kg}
\end{figure*}

A domain-specific biomedical graph was constructed to support intervention-oriented reasoning in cardiovascular care (Figure~\ref{fig:kg}). Nodes represent diseases, drugs, non-drug interventions, protein targets, biological processes, risk factors, clinical outcomes, adverse effects, contraindications, populations, evidence sources, studies, and reified assertions. The final entity table contains 117 biomedical and contextual nodes, comprising 10 diseases, 15 drugs, 10 interventions, 15 protein targets, 13 biological processes, 13 risk factors, 10 clinical outcomes, 14 adverse effects, 8 contraindications, and 9 populations. The graph was synthesized from the underlying biomedical evidence and subsequently reviewed by two medical annotators to assess the accuracy and consistency of the entity annotations.

\subsubsection{Assertion reification: every claim as a first-class node}\label{subsubsec:reification}

\begin{figure*}[t]
\centering
\includegraphics[width=0.9\textwidth]{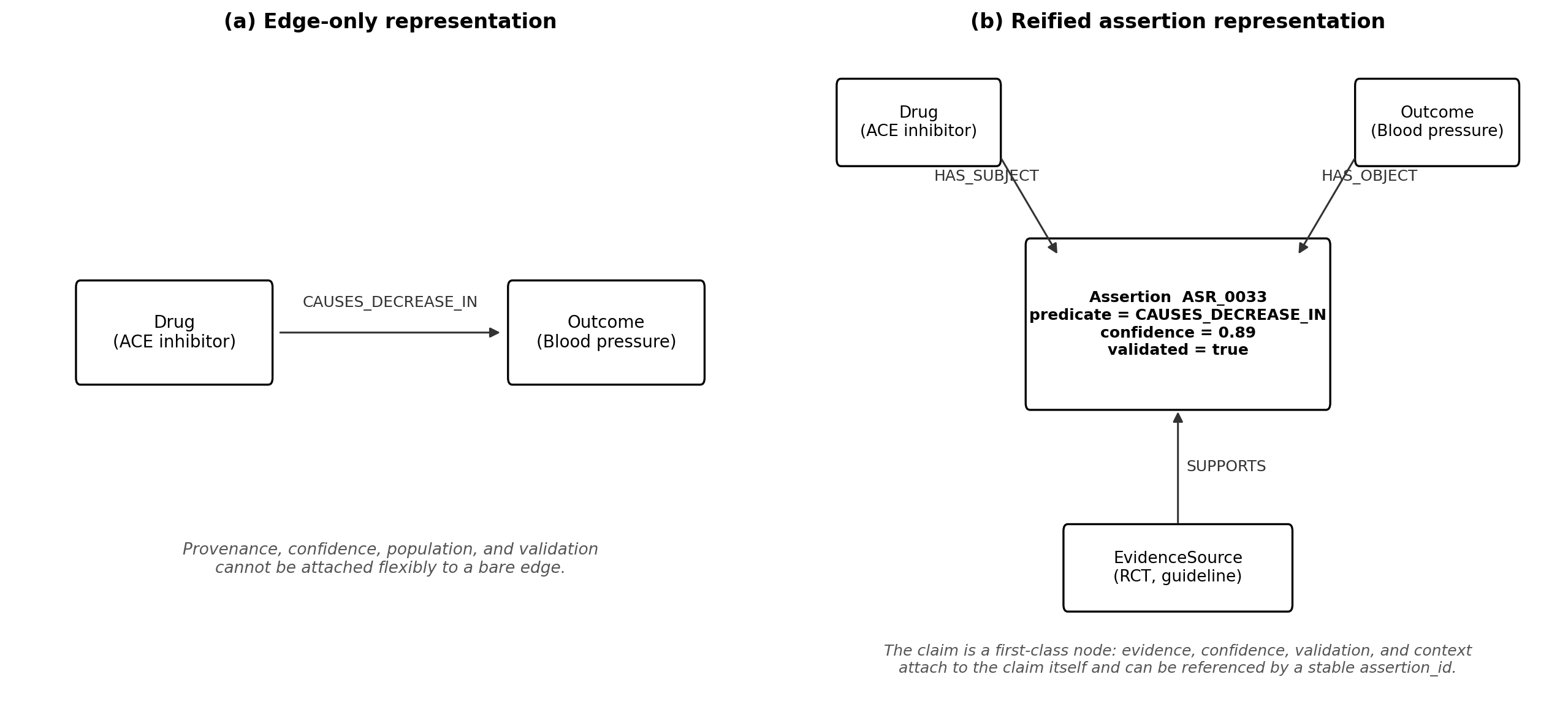}
\caption{Assertion reification schema. (a)~An edge-only representation of a biomedical claim, in which provenance, confidence, population, and validation cannot be attached flexibly. (b)~Reified representation: the claim is a first-class Assertion node with a stable identifier, linked to its subject and object through \texttt{HAS\_SUBJECT} and \texttt{HAS\_OBJECT}, and to one or more Evidence Source nodes through \texttt{SUPPORTS}.}\label{fig:reification}
\end{figure*}

The central design decision is that every biomedical relationship is represented not only as a direct Neo4j edge but also as a reified assertion node. In a conventional property-graph encoding, a claim such as ``an ACE inhibitor decreases blood pressure'' would be stored only as an edge, \texttt{(Drug)-[:CAUSES\_DECREASE\_IN]->(Outcome)}. In our graph, the same claim is additionally stored as a dedicated Assertion node with a stable identifier (e.g., \texttt{ASR\_0033}) linked to its endpoints through explicit \texttt{HAS\_SUBJECT} and \texttt{HAS\_OBJECT} relationships:

\begin{verbatim}
(:Assertion {assertion_id: "ASR_0033"})
(:Assertion)-[:HAS_SUBJECT]->(:Drug)
(:Assertion)-[:HAS_OBJECT]->(:Outcome)
\end{verbatim}

Each assertion node carries structured properties: \texttt{predicate}, \texttt{relation\_class}, \texttt{direction}, \texttt{causal\_status}, \texttt{evidence\_level}, \texttt{confidence}, \texttt{validated}, \texttt{context}, \texttt{population\_id}, \texttt{temporality}, and a \texttt{source\_databases\_json} field. These attributes attach naturally to a node as shown in Figure~\ref{fig:reification}.

\begin{figure*}[t]
\centering
\includegraphics[width=0.9\textwidth]{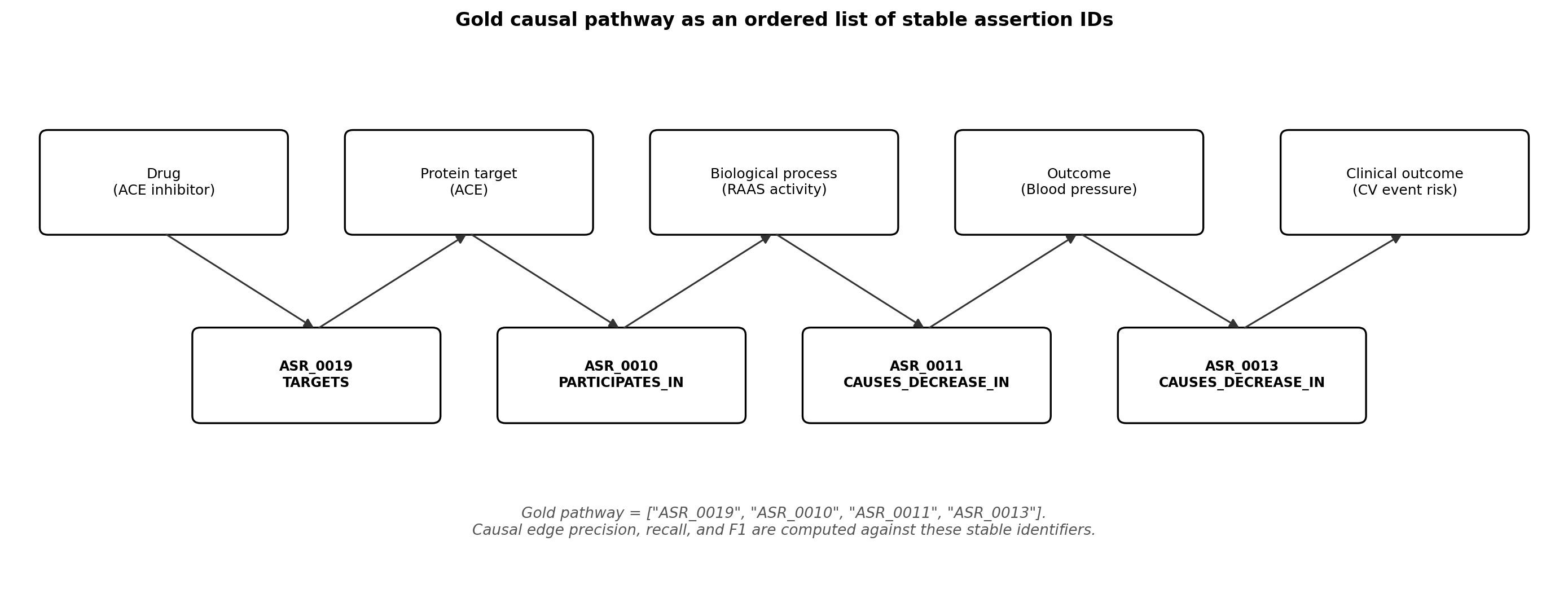}
\caption{Worked example of a gold causal pathway. Each biomedical entity is connected to the next through a reified assertion node with a stable identifier, so that a gold pathway is expressible as an ordered list of \texttt{assertion\_id}s and consumed directly by the scoring metrics for causal edge precision, recall, and F1.}\label{fig:gold-path}
\end{figure*}

Reification is used for four reasons that are directly consequential for this study. First, a biomedical claim is not fully characterised by the pair of entities it connects; it also has a causal status, a confidence, a validation state, an applicable population, and a temporal scope. These live on the claim itself rather than being duplicated across parallel edges. Second, evidence links attach to a specific claim rather than to an opaque edge, expressing patterns such as \texttt{(:EvidenceSource)-[:SUPPORTS]->(:Assertion)}. Third, when several sources back the same relationship, an edge-only model tends to accumulate near-duplicate edges; reification preserves a single canonical claim node and lets many evidence links point to it, keeping the graph compact and provenance auditable. Fourth, and most consequential for evaluation, stable \texttt{assertion\_id}s make gold structures precise and referable: a gold pathway can be expressed as an ordered list such as \texttt{["ASR\_0019", "ASR\_0010", "ASR\_0011", "ASR\_0013"]} as shown in Figure~\ref{fig:gold-path}, which is directly usable by the scoring pipeline of Section~\ref{subsec:scoring}.

Direct typed edges (\texttt{CAUSES\_DECREASE\_IN}, \texttt{TARGETS}, \texttt{MAY\_CAUSE\_ADVERSE\_EFFECT}, and others) are retained alongside the reified assertions. This dual representation is deliberate: direct edges support fast traversal and simple Cypher retrieval for the C2, C3, and C4 grounding conditions, while assertion nodes carry provenance, validation, and gold-path referencing. Every direct edge in the graph is paired with a corresponding assertion node that shares the same subject, predicate, and object.

The expanded graph contains 257 assertions and 15 evidence sources, including 11 study-level records. Direct biomedical relationships are preserved for traversal efficiency while remaining linked to assertion objects through stable identifiers. Relationship totals include 15 \texttt{TARGETS} edges, 14 \texttt{PARTICIPATES\_IN} edges, 65 \texttt{CAUSES\_INCREASE\_IN} edges, 82 \texttt{CAUSES\_DECREASE\_IN} edges, 18 \texttt{ASSOCIATED\_WITH} edges, 15 \texttt{INDICATED\_FOR} edges, 26 \texttt{MAY\_CAUSE\_ADVERSE\_EFFECT} edges, 13 \texttt{CONTRAINDICATED\_FOR} edges, and 9 \texttt{MODIFIES\_EFFECT\_OF} edges. Together with the reified layer, this representation supports efficacy, safety, contextual applicability, and evidence provenance, each addressable independently by the scoring pipeline.

\subsection{Grounding conditions}\label{subsec:grounding}

Four grounding conditions determine how the extracted subgraph is composed into the model's context.

\begin{itemize}
\item \textbf{C1 (ungrounded).} Only the scenario and intervention task.
\item \textbf{C2 (knowledge-graph).} Scenario plus knowledge-graph context and evidence provenance, without full causal-path emphasis.
\item \textbf{C3 (causal-graph).} Scenario plus validated causal pathways, with limited contextual evidence detail.
\item \textbf{C4 (integrated).} Scenario plus integrated causal structure, mechanistic context, evidence provenance, contraindications, adverse-effect information, and uncertainty-relevant context.
\end{itemize}

The conditions vary the type of retrieved context rather than its volume, so that observed differences can be attributed to context organisation rather than to prompt style. C1 bypasses subgraph extraction by design; it is the ungrounded baseline.

\subsection{Scenario-conditioned subgraph extraction and prompt execution}\label{subsec:extraction}

Given a clinical scenario, framework component 2 retrieves the relevant reified-assertion subgraph from Neo4j using parameterized Cypher queries prepared separately for C2, C3, and C4. The prompt itself is then constructed deterministically from the extracted subgraph; the same scenario header, candidate options, output schema, and JSON contract are shared by all four conditions, and only the \emph{Grounded context} block changes. C1 bypasses extraction by design and receives no graph content. This isolation is what allows behavioral differences across conditions to be attributed to context organization rather than to prompt wording, task framing, or output schema. The same extraction interface applies unchanged to a scenario supplied from any external source (a clinician-entered vignette, an EHR-derived summary, or an external benchmark item).

To make the mechanism concrete, we walk through a single scenario, SCN\_0001 (a direct-effect item: adult with persistent hypertension, target outcome $=$ systolic blood pressure, candidate options $=$ lisinopril, atenolol, smoking cessation), and show what each condition receives in the \emph{Grounded context} block of the generated prompt. Content outside that block is identical across the four conditions and is omitted for brevity.

\paragraph{C1 (ungrounded).} No subgraph supplied in the prompt context section. The model sees only the scenario, the candidate options, and the JSON output schema. There is no assertion identifier space to reference, so the model cannot populate \texttt{causal\_path\_assertion\_ids} or \texttt{supporting\_evidence\_ids} truthfully; it must either leave them empty or invent identifiers. This baseline exists to measure what happens when the panel has no graph to score against.

\paragraph{C2 (knowledge-graph context).} The extractor returns candidate-level indication, association, and adverse-effect assertions with evidence identifiers, but no mechanistic chain. C2 gives the model the knowledge needed to justify a choice (indications, associations, one adverse effect) but not the causal decomposition of how the intervention reaches the outcome.

\paragraph{C3 (causal-graph context).} The extractor returns the validated causal chain of \texttt{assertion\_id}s from intervention to outcome (target $\rightarrow$ biological process $\rightarrow$ intermediate outcome $\rightarrow$ clinical outcome), plus a directly attached adverse-effect assertion. C3 exposes the reified mechanism directly. The gold causal pathway for scoring is exactly \texttt{["ASR\_0019", "ASR\_0010", "ASR\_0011", "ASR\_0013"]}, so a correct C3 response can populate \texttt{causal\_path\_assertion\_ids} verbatim.

\paragraph{C4 (integrated causal-knowledge-graph context).} The extractor returns the C3 causal chain, but each assertion is annotated with its evidential and provenance status (guideline vs.\ validated causal edge, evidence tier, source-database attribution), and the surrounding treatment and safety context is retained. C4 supplies the same underlying \texttt{assertion\_id}s as C3 but presents them alongside the knowledge-graph indication (\texttt{ASR\_0028}) and adds a typed evidential label to each row (guideline, validated causal edge with evidence tier and source, safety evidence). This is the condition designed to test whether a model can not only reconstruct the mechanism but also weigh evidence quality and preserve provenance in its cited support.

Across the four conditions, three properties are preserved by construction. First, the scenario, candidate options, and JSON output schema are byte-identical; only the \emph{Grounded context} block varies. Second, every assertion identifier in every condition resolves to a node in the same reified-assertion graph, so the scoring pipeline (Section~\ref{subsec:scoring}) computes causal edge precision/recall/F1, evidence accuracy, and unsupported-claim rate against the same reference regardless of which condition produced the output. Third, C1's absence of grounded context is not a bug in the extraction step but the intended baseline: it produces the row of the metric panel that tells us what a model does when it has no graph to be measured against.

\subsection{Evaluation methodology: repeated-measures protocol}\label{subsec:protocol}

Each scenario is presented under all four grounding conditions in the same run, so within-scenario condition deltas can be computed. For a single model, the full benchmark comprises 320 runs (80 scenarios $\times$ 4 conditions). This protocol is an evaluation instrument used to demonstrate the framework; it is not itself a component of the framework.

\subsection{Evaluation instrument: category-balanced scenario generator}\label{subsec:generator}

The framework itself accepts any clinical scenario as input. To evaluate the framework in a controlled way, however, we need input scenarios whose gold causal, safety, and evidential structure is known in advance, so that the scoring pipeline (Section~\ref{subsec:scoring}) can operate against defined targets. We therefore developed a category-balanced scenario generator that produces 80 structured intervention items directly from the knowledge graph.

The generated scenarios are evenly balanced across eight reasoning categories, each chosen to probe a distinct causal-reasoning failure mode: direct effect, mediated effect, multiple pathways, confounded association, competing interventions, contraindication-sensitive reasoning, adverse-effect pathways, and incomplete or uncertain evidence. Each scenario defines the clinical context, the decision target, the candidate actions, the expected gold intervention, and the gold causal and evidential structure expressed as ordered lists of stable \texttt{assertion\_id}s. Category balance is enforced so that no single class of clinical reasoning problem can dominate an aggregate metric, and so that category-level dissociations across the metric panel become observable. The generator is an evaluation instrument, not a framework component; in downstream deployment it is replaced by scenarios drawn from clinicians, EHR summaries, or external benchmarks.

\subsection{Automated scoring and evaluation pipeline}\label{subsec:scoring}

Model outputs are not scored as free text. Each response is first parsed into a fixed set of structured fields---the recommended intervention, the reasoning pathway, the adverse effects surfaced, the contraindications noted, the evidence sources cited, and any explicit statement about evidential sufficiency. These fields are then compared, on a single scoring pass, against gold structures that are themselves expressed as sets of \texttt{assertion\_id}s drawn from the same reified-assertion graph used for grounding. Anchoring both sides of the comparison on the same identifier space is what makes the pipeline reproducible: no free-text matching against gold pathways is required for the causal, adverse-effect, contraindication, or evidence metrics.

The primary outcome is \emph{intervention accuracy}, and its stricter companion \emph{preferred intervention accuracy}. Intervention accuracy asks whether the model selected an appropriate action from the candidate set defined by the scenario; preferred intervention accuracy asks whether it selected the specific gold intervention marked as preferred when the scenario admits more than one clinically acceptable choice. This distinction is deliberate: in a decision-support setting it is possible to be ``not wrong'' (any of several acceptable actions) without being ``right'' (the action the guidelines actually prefer for that context), and separating the two metrics lets the framework detect that gap rather than collapse it into one number.

\emph{Causal edge precision, recall, and F1} score how faithfully the model reconstructed the mechanism that justifies its recommendation. The scenario carries a gold causal pathway, expressed as an ordered set of \texttt{assertion\_id}s that link the intervention through its molecular target, biological process, and clinical outcome. Precision asks what fraction of the causal edges the model stated are present in the gold set; recall asks what fraction of the gold edges the model surfaced; F1 combines the two. Together they distinguish a model that arrives at the correct action by inventing a plausible-sounding but unsupported mechanism (high accuracy, low precision) from one that arrives at the correct action through the mechanism the graph sanctions (high accuracy, high precision and recall).

\emph{Adverse-effect F1 and contraindication recall} score the safety axis. Adverse-effect F1 is computed only on scenarios that have a gold adverse-effect target and asks whether the model surfaced the specific \texttt{MAY\_CAUSE\_ADVERSE\_EFFECT} claims that apply to the recommended intervention. Contraindication recall asks, on contraindication-sensitive scenarios, whether the model surfaced the contextual constraint that should have blocked (or preferred against) a competing option---for example, avoiding an ACE inhibitor in a pregnant patient. Reporting these separately from intervention accuracy is important because a model can pick the correct action while omitting the harm or the contraindication that made it the correct action; a benchmark that only checked the final action would score that response identically to a fully justified one.

\emph{Evidence accuracy and unsupported claim rate} score the evidential axis. Evidence accuracy asks whether the sources the model cited are in fact the \texttt{EvidenceSource} nodes that \texttt{SUPPORTS} the assertions the model relied on; a citation to an existing source is not enough---it must actually support the claim being made. Unsupported claim rate (lower is better) asks the complementary question: what fraction of the causal or mechanistic claims produced by the model do not have a corresponding subject--predicate--object pattern in the retrieved graph context. Together, these two metrics separate ``the model cited its sources correctly'' from ``the model made claims the graph did not sanction,'' which are different failure modes and require different fixes.

\emph{Uncertainty correctness} scores whether the model recognized that the evidence was insufficient when the scenario was designed to test exactly that. On the incomplete-evidence category, the gold behavior is not a specific action but an explicit acknowledgment that the available evidence does not support a confident recommendation. Measuring this separately is what prevents the benchmark from silently rewarding a confident-sounding wrong answer over a correctly hedged one, and it is what let the pilot detect the dissociation between recognizing insufficiency (uncertainty correctness 1.000) and producing the gold intervention decision (intervention accuracy 0.000) in that category.

Because all seven metrics are computed on the same scoring pass, against the same graph, and against the same gold \texttt{assertion\_id} sets, they can be read as one coherent panel rather than as seven independent evaluations. This is the property the framework relies on downstream: when C1 and C4 differ on intervention accuracy but differ in the opposite direction on causal F1 and unsupported claim rate, that pattern is directly interpretable because both numbers came from the same run against the same reference. A benchmark that assembled these metrics from separate scoring passes could not make that inference cleanly.

Model outputs are parsed into structured response fields and scored against gold structures expressed as sets of \texttt{assertion\_id}s; a worked example is given in Table~\ref{tab:worked-example}. The primary outcomes are intervention accuracy and preferred intervention accuracy. Additional outcomes are causal edge precision, causal edge recall, causal edge F1, adverse-effect F1, contraindication recall, evidence accuracy, unsupported claim rate (lower is better), and uncertainty correctness. These outcomes are selected to expose multiple dimensions of intervention-oriented reasoning on the same scoring pass: whether the model chose an appropriate action, whether it justified that action with defensible causal structure, whether it surfaced harms and contraindications, whether it cited support accurately, whether it hallucinated unsupported mechanisms, and whether it correctly recognized insufficiency of evidence.

\begin{table*}[t]
\caption{Worked example of one scenario passing through the scoring pipeline. The scenario is drawn from the direct-effect category; identifiers (\texttt{ASR\_*}, \texttt{EV\_*}) refer to nodes in the reified-assertion graph. The example is illustrative and does not affect the aggregate numbers in Section~\ref{sec:results}.}\label{tab:worked-example}%
\begin{tabular}{@{}p{0.24\textwidth}p{0.70\textwidth}@{}}
\toprule
Stage & Content \\
\midrule
Scenario input & 62-year-old with essential hypertension, LDL 138\,mg/dL, no diabetes, eGFR 78, not pregnant. Decision target: choose the preferred first-line antihypertensive from \{ACE inhibitor, thiazide diuretic, beta-blocker, calcium-channel blocker\}. \\
Gold intervention & ACE inhibitor (lisinopril). \\
Gold causal pathway & \texttt{[ASR\_0019, ASR\_0010, ASR\_0011, ASR\_0013]} $=$ ACE inhibitor \texttt{TARGETS} ACE $\rightarrow$ \texttt{PARTICIPATES\_IN} RAAS activity $\rightarrow$ \texttt{CAUSES\_DECREASE\_IN} blood pressure $\rightarrow$ \texttt{CAUSES\_DECREASE\_IN} cardiovascular event risk. \\
Gold adverse effects & \{\texttt{ASR\_0087} (dry cough), \texttt{ASR\_0091} (hyperkalemia)\}. \\
Gold contraindication scope & Nonapplicable in this scenario (patient is not pregnant); metric is undefined. \\
Gold evidence sources & \{\texttt{EV\_0007} (JNC-8 guideline), \texttt{EV\_0011} (HOPE trial)\}. \\
Expected uncertainty behavior & Confident recommendation (evidence is sufficient). \\
Raw model output (abridged) & ``I recommend lisinopril. It blocks angiotensin II production and thereby lowers blood pressure, which reduces cardiovascular events. It also lowers LDL cholesterol. The main adverse effect to counsel is a dry cough. Supporting evidence: HOPE trial.'' \\
Parsed structured fields & \texttt{intervention = "lisinopril"}; \texttt{causal\_pathway\_ids = [ASR\_0019, ASR\_0011, ASR\_0013]}; \texttt{adverse\_effect\_ids = [ASR\_0087]}; \texttt{contraindication\_ids = []}; \texttt{evidence\_source\_ids = [EV\_0011]}; \texttt{unsupported\_free\_text\_claims = ["lowers LDL cholesterol"]}; \texttt{uncertainty\_flag = false}. \\
Scoring against gold & intervention accuracy $=$ 1.00 (lisinopril $\in$ ACE-inhibitor class); preferred intervention accuracy $=$ 1.00; causal edge precision $=$ 3/3 $=$ 1.00; causal edge recall $=$ 3/4 $=$ 0.75 (missed \texttt{ASR\_0010}, the RAAS mediator); causal edge F1 $=$ 0.857; adverse-effect F1 $=$ 0.667 (surfaced \texttt{ASR\_0087}, missed \texttt{ASR\_0091}); contraindication recall $=$ NA; evidence accuracy $=$ 1/1 $=$ 1.00 (\texttt{EV\_0011} supports the cited pathway); unsupported claim rate $=$ 1/4 $=$ 0.25 (the LDL claim has no matching S--P--O in the retrieved subgraph); uncertainty correctness $=$ 1.00 (no hedge expected, none produced). \\
\botrule
\end{tabular}
\end{table*}

All seven metrics fall out of one parse-then-compare pass over the same output. The model would score 1.00 on intervention accuracy alone, but the panel simultaneously reveals a missed mediator (recall gap), a missed adverse effect (safety gap), and a fabricated causal claim (unsupported-claim signal). These are the failure modes that a single-endpoint accuracy report would silently hide, and they are surfaced here because every metric is anchored on the same \texttt{assertion\_id} reference.

\subsection{Experimental setup}\label{subsec:experimental-setup}

The framework and instruments were implemented as Python scripts within a single workspace. The graph substrate uses Neo4j, populated from CSV source tables (\texttt{entities.csv}, \texttt{assertions.csv}, \texttt{evidence\_sources.csv}, \texttt{assertion\_evidence.csv}, and related) via a bulk import script and Cypher validation queries. The model under test in the pilot is gpt-5.4, invoked through the OpenAI Batch API to run the full 320-prompt matrix in one job. Prompts are constructed deterministically from the retrieved subgraph; there is no per-scenario prompt engineering. Scoring is fully automatic and is anchored to \texttt{assertion\_id}s so that no free-text matching against gold structures is required for the causal, adverse-effect, contraindication, or evidence metrics. Of the 320 planned runs, 313 were successfully scored; the remaining seven were skipped because of missing model responses associated with API failure or incomplete output capture, yielding condition-specific counts of 77 (C1), 79 (C2), 78 (C3), and 79 (C4).

\section{Results}\label{sec:results}

The metric panel separates the four conditions along interpretable, non-redundant axes, which is the property the framework is intended to have. Table~\ref{tab:panel} summarises the overall panel. The $n$ column is the number of scored scenarios for which the metric is defined (for example, causal edge precision is undefined for C1 because no causal edges are produced without grounding; adverse-effect F1 is defined only on scenarios with a gold adverse-effect target).

\begin{table*}[t]
\caption{Overall metric panel across grounding conditions in the cardiovascular pilot demonstration (gpt-5.4). NA indicates the metric is undefined for that condition given the scoring rule. Lower is better for unsupported claim rate; higher is better for all other metrics.}\label{tab:panel}%
\begin{tabular*}{\textwidth}{@{\extracolsep\fill}lcccc@{}}
\toprule
Metric & C1 (mean, $n$) & C2 (mean, $n$) & C3 (mean, $n$) & C4 (mean, $n$) \\
\midrule
Intervention accuracy               & 0.948, 77 & 0.873, 79 & 0.808, 78 & 0.886, 79 \\
Preferred intervention accuracy     & 0.948, 77 & 0.873, 79 & 0.808, 78 & 0.886, 79 \\
Causal edge precision               & NA, 0     & 0.522, 79 & 0.808, 78 & 0.806, 79 \\
Causal edge recall                  & 0.000, 77 & 0.785, 79 & 0.808, 78 & 0.886, 79 \\
Causal edge F1                      & NA, 0     & 0.579, 79 & 0.808, 78 & 0.838, 79 \\
Adverse-effect F1                   & NA, 0     & 0.733, 20 & 0.822, 30 & 0.833, 40 \\
Contraindication recall             & 1.000, 10 & 1.000, 10 & 1.000, 10 & 0.800, 10 \\
Evidence accuracy                   & NA, 0     & 0.705, 79 & 0.679, 78 & 0.738, 79 \\
Unsupported claim rate ($\downarrow$ better) & NA, 0 & 0.259, 79 & 0.192, 78 & 0.114, 79 \\
Uncertainty correctness (overall)   & 0.130, 77 & 0.114, 79 & 0.115, 78 & 0.114, 79 \\
\botrule
\end{tabular*}
\end{table*}

The ungrounded baseline C1 obtains the highest raw intervention accuracy (0.948). The grounded conditions produce lower raw intervention accuracy (0.873 for C2, 0.808 for C3, 0.886 for C4). On the causal and evidential axes, the ordering is different: C4 achieves the highest causal edge F1 (0.838), the highest adverse-effect F1 (0.833), the highest evidence accuracy (0.738), and the lowest unsupported claim rate (0.114). We report these numbers as evidence that the multi-metric panel exposes behavior that a single-endpoint accuracy report would hide: on this pilot, ranking the conditions by intervention accuracy alone would have placed C1 first, but that ordering carries no information about mechanism fidelity, evidence use, or unsupported claim generation, all of which the framework measures separately. The conditions are not collinear on the panel. We do not draw a clinical-performance conclusion from the ranking itself.

\subsection{Category-level behavior of the metric panel}\label{subsec:category-behavior}

The scenario categories produced distinct metric profiles, supporting the framework's design premise that intervention reasoning is not a single competency and should not be measured with a single number. In multiple-pathway scenarios, C4 recorded perfect intervention accuracy and causal recall together with evidence accuracy of 1.000 and an unsupported claim rate of 0.000. In adverse-effect pathway scenarios, C4 recorded perfect intervention accuracy, perfect causal precision and recall, and adverse-effect F1 of 1.000. In contraindication-sensitive scenarios, C4 maintained perfect intervention accuracy and perfect causal pathway recovery. These are useful as evidence that the framework's category structure interacts meaningfully with the grounding manipulation; they are not clinical-performance guarantees.

On the incomplete or uncertain evidence category, all three grounded conditions scored 0.000 on intervention accuracy while uncertainty correctness was 1.000 across all conditions. This dissociation between correctly recognising evidential insufficiency and producing the gold intervention decision is precisely the kind of finding the multi-metric panel is designed to expose, and it identifies an explicit benchmark-design choice for the community: whether abstention and uncertainty acknowledgment should be scored as separate targets from final action selection.

C3 showed a distinct profile on multiple-pathway scenarios (intervention accuracy 0.400, causal precision 0.400, causal recall 0.400, evidence accuracy 0.400, unsupported claim rate 0.600). We report this as evidence that the framework separates causal-only and integrated-causal grounding along interpretable axes rather than collapsing them; whether the pattern generalises across models is a claim reserved for the multi-model phase.

\subsection{Example paired-condition contrast (C4 vs.\ C2)}\label{subsec:paired-contrast}

To show how the framework's paired within-scenario deltas behave, we examined C4 relative to C2 (Table~\ref{tab:paired}). Relative to C2, C4 shows a small positive delta on intervention accuracy ($+0.013$), a larger delta on causal precision ($+0.284$), positive deltas on causal recall ($+0.101$), causal F1 ($+0.259$), adverse-effect F1 ($+0.100$), and evidence accuracy ($+0.033$), and a favorable negative delta on unsupported claim rate ($-0.145$). Contraindication recall is 0.200 lower in C4; the diagnostic in Section~\ref{subsec:diagnostic} attributes this to output-schema semantics in a small number of scenarios rather than to missing safety knowledge in the graph. The purpose of reporting this contrast is to demonstrate that the paired-delta layer of the framework operates end-to-end and yields metric-specific rather than lumped signal. It is not a claim that integrated grounding is superior to knowledge-graph grounding for clinical use; that claim would require the multi-model phase.

\begin{table*}[t]
\caption{Illustrative within-scenario paired deltas for the C4 vs.\ C2 contrast in the pilot. Positive deltas indicate improvement for C4 on the metric; for unsupported claim rate, a negative delta is favorable.}\label{tab:paired}%
\begin{tabular*}{\textwidth}{@{\extracolsep\fill}lcccc@{}}
\toprule
Metric & C2 mean & C4 mean & $\Delta$ (C4$-$C2) & Direction favorable to C4? \\
\midrule
Intervention accuracy   & 0.873 & 0.886 & $+0.013$ & Yes \\
Causal edge precision   & 0.522 & 0.806 & $+0.284$ & Yes \\
Causal edge recall      & 0.785 & 0.886 & $+0.101$ & Yes \\
Causal edge F1          & 0.579 & 0.838 & $+0.259$ & Yes \\
Adverse-effect F1       & 0.733 & 0.833 & $+0.100$ & Yes \\
Evidence accuracy       & 0.705 & 0.738 & $+0.033$ & Yes \\
Unsupported claim rate  & 0.259 & 0.114 & $-0.145$ & Yes (lower is better) \\
Contraindication recall & 1.000 & 0.800 & $-0.200$ & No (see Section~\ref{subsec:diagnostic}) \\
\botrule
\end{tabular*}
\end{table*}

\subsection{Diagnostic value of the framework}\label{subsec:diagnostic}

Two diagnostic observations illustrate that the framework surfaces evaluation-design questions in addition to model behavior. First, the high intervention accuracy of C1 was partly attributable to the incomplete or uncertain evidence category, where the grounded conditions scored 0.000 on intervention accuracy despite scoring 1.000 on uncertainty correctness. Without a benchmark that measures uncertainty correctness separately, an ungrounded model could appear superior on a single accuracy endpoint even when the gold behavior was abstention or uncertainty-calibrated reasoning. The framework makes that failure mode explicit and forces the benchmark designer to choose whether to reward the recognised-uncertainty response.

Second, the lower contraindication recall observed in C4 was traced to two scenarios in which the model recommended the correct intervention (amlodipine) but returned an empty contraindication field even though pregnancy was expected as the gold contraindication context. Inspection of the grounded prompts confirmed that the relevant contraindication information was present in the retrieved context. The most likely explanation is that the model interpreted the contraindication output field narrowly (contraindications of the selected intervention), whereas the scorer interpreted it broadly (a scenario-level contraindication context that helped rule out competing options). We report this because it illustrates that the framework detects output-schema-driven measurement artefacts distinct from missing knowledge, which is a class of failure that a coarser benchmark could not distinguish from a genuine safety gap.

\section{Discussion}\label{sec:discussion}

The principal contribution of this manuscript is an evaluation framework, not a clinical-performance claim. The framework has four components: a provenance-preserving causal knowledge graph with reified assertions, a scenario-conditioned subgraph extraction step, a controlled set of grounding conditions that vary the type of retrieved context, and a multi-dimensional automatic scoring pipeline anchored to stable assertion identifiers. Together they define how a claim moves from the underlying knowledge base into a model output that can be scored on causal, safety, and evidential axes. To test the framework in a controlled way, we additionally develop a category-balanced scenario generator that supplies structured input items with graph-anchored gold answers, and a repeated-measures design that presents each scenario under all four conditions; both are evaluation instruments, not framework components. The pilot demonstrates that this pipeline is implementable end-to-end on a single workspace and that its metric panel is discriminative across the conditions and scenario categories it is designed to distinguish. Reporting only intervention accuracy would have hidden most of the between-condition behavior we observe, including differences in causal reconstruction, evidence use, adverse-effect recovery, and unsupported-claim generation. This is the property a benchmark for intervention-oriented healthcare LLMs needs, independent of which model is under test.

Because the pilot instantiates the framework on a single model, it does not, and is not intended to, establish that any grounding condition is clinically superior. A common alternative reading of the descriptive numbers would take C1's high intervention accuracy as evidence that ungrounded models are sufficient. The framework itself is the response to that reading: the causal, evidential, adverse-effect, contraindication, and unsupported-claim axes are measured explicitly precisely so that a single-endpoint answer cannot silently substitute for decision-relevant reasoning.

The framework is positioned within the broader shift from predictive to causal health AI. In that setting, evaluation infrastructure needs to detect whether a generative system connects its recommendations to explicit causal assumptions, supporting evidence, harms, and uncertainty, rather than whether a single answer matches a gold label. The framework operationalises that requirement by asking, for each scenario, not only which intervention was chosen but also which causal edges the model reconstructed, which adverse effects and contraindications it surfaced, which evidence it cited, and where it produced claims unsupported by the retrieved graph. Because the type of retrieved context is a manipulable variable, the framework can in principle discriminate models and prompting strategies that are sensitive to context organisation from those that are sensitive only to context volume.

Three implications follow. First, benchmarks for healthcare LLMs should treat final-answer accuracy as one axis among several, and should measure unsupported claims, evidence alignment, harms, contraindications, and uncertainty as first-class outcomes. Second, structured biomedical graphs are usefully treated not only as retrieval resources but as experimental instruments whose contents and organisation can be varied to test whether a model responds to the type of structure supplied. Third, output-schema design is a component of benchmark validity, not a peripheral engineering detail: the contraindication diagnostic showed that a model may possess the relevant safety context yet fail to communicate it in the expected form, and a benchmark that cannot separate this from a genuine knowledge gap is measuring the wrong thing. The framework also exposes a task-design question for uncertainty-aware causal health AI. In the incomplete-evidence category the grounded conditions recognised insufficiency correctly (uncertainty correctness 1.000) yet failed to produce the gold intervention decision. This is not primarily a language-generation failure; it is a decision-policy and scoring-design choice about whether abstention or explicit ``insufficient evidence'' outputs should count as correct interventions. The framework makes that choice explicit and therefore auditable.

\subsection{Limitations}\label{subsec:limitations}

Several limitations follow directly from the framing of this work as a benchmark and methodology contribution. First, the empirical component is a single-model pilot; it demonstrates that the framework is implementable and discriminative, not clinical performance for any model. Comparative model claims and effect-size estimates are deferred to the planned multi-model repeated-measures phase. Second, the benchmark is a designed evaluation environment in a cardiovascular sub-domain and does not, on its own, establish external validity for live clinical deployment or for other clinical areas; the framework is intended to be reinstantiated per domain.

\subsection{Future work}\label{subsec:future-work}

Future work will focus on two directions. First, the framework will be evaluated across a broader set of frontier and open-weight LLMs using a repeated-measures design. Second, cross-domain instantiation of the framework beyond cardiovascular care will be pursued to test whether the category structure and metric panel transfer.

\section{Conclusion}\label{sec:conclusion}

We propose a graph-centered grounding framework for evaluating intervention-oriented LLM behavior in healthcare. The framework integrates four components: a provenance-preserving causal knowledge graph with reified assertions, scenario-conditioned subgraph extraction, controlled grounding conditions that vary the retrieved context, and a multidimensional automatic scoring pipeline linked to stable \texttt{assertion\_id}s. We also develop a category-balanced scenario generator that produces structured evaluation cases with graph-anchored gold answers across eight causal-reasoning failure modes. The generator serves as an evaluation instrument rather than a component of the framework.

A cardiovascular pilot comprising 313 scored runs with GPT-5.4 demonstrates that the framework can be implemented end to end and that its evaluation metrics distinguish performance across grounding conditions and scenario categories. The results also show the value of multidimensional evaluation by revealing patterns that would be obscured by a single accuracy measure, including the ability to distinguish output-schema misalignment from substantive knowledge gaps. The pilot is therefore intended to demonstrate the interpretability and usefulness of the evaluation framework rather than to establish definitive clinical performance for a particular model. Comparative conclusions across models are reserved for the planned multi-model repeated-measures study, for which the proposed framework provides the underlying evaluation infrastructure.

\backmatter

\bmhead{Supplementary information}

Not applicable.

\bmhead{Acknowledgements}

Not applicable.

\section*{Declarations}

\bmhead{Funding}
No funding was received to assist with the preparation of this manuscript. [Revise if applicable and include grant number(s).]

\bmhead{Competing interests}
The authors have no relevant financial or non-financial interests to disclose. [Revise if applicable.]

\bmhead{Ethics approval}
This study did not involve human participants, human data, or human biological material. The benchmark consists of synthetic clinical scenarios generated from a curated biomedical knowledge graph and evaluated against automated scoring metrics; no patient-level data were collected or analysed. Institutional review board approval was therefore not required.

\bmhead{Consent to participate and consent to publish}
Not applicable. This work does not include data from identifiable individuals.

\bmhead{Data availability}
The benchmark source tables (entity, assertion, evidence-source, and scenario tables), the Neo4j import files, and the aggregated per-condition results tables that support the findings of this study are available on request to the corresponding author.

\bmhead{Code availability}
The graph construction, prompt generation, retrieval, experiment execution, automated scoring, and analysis scripts used in this study are provided in the same companion repository under an open-source license [license to be added upon acceptance].

\bmhead{Author contributions}
All authors contributed to the study conception and design. [Author One] led the framework design, graph construction, and manuscript drafting. [Authors Two and Three] developed the scoring pipeline and analytic summaries. [Author Four] contributed to scenario generation and reviewed the manuscript.

\bmhead{Declaration on Generative AI}
An AI assistant was used for AI-assisted copy editing (readability, grammar, and formatting) of the manuscript text. The authors reviewed and edited the content as needed and take full responsibility for the publication's content.

\bibliography{jhir-bibliography}

\end{document}